\documentclass[times,3p]{elsarticle}
\usepackage[english]{babel}     
\usepackage[utf8]{inputenc}   
\usepackage{amsmath}            
\usepackage{epstopdf}           
\usepackage{flushend,xcolor,comment}

\usepackage{epstopdf}

\usepackage{amssymb}

\usepackage[figuresright]{rotating}

\begin{document}
	
	\begin{frontmatter}
		\title{Machine learning to assess relatedness: the advantage of using firm-level data}
		
		\author[1,2]{G. Albora\corref{cor1}}
		\ead{alboragiambattista@gmail.com}
		
		\author[3,1]{A. Zaccaria}

		\cortext[cor1]{Corresponding Author.}
		
		\address[1]{Enrico Fermi Center for Study and Research, Rome, Italy.}
		\address[2]{Sapienza University, Department of Physics, Rome, Italy }
		\address[3]{Institute for Complex Systems - CNR, UOS Sapienza, Rome, Italy.}
		
		\begin{abstract}
			The relatedness between a country or a firm and a product is a measure of the feasibility of that economic activity.  As such, it is a driver for investments at a private and institutional level.  Traditionally, relatedness is measured using networks derived by country-level co-occurrences of product pairs, that is counting how many countries export both. In this work, we compare networks and machine learning algorithms trained not only on country-level data, but also on firms, that is something not much studied due to the low availability of firm-level data. We quantitatively compare the different measures of relatedness, by using them to forecast the exports at the country and firm-level, assuming that more related products have a higher likelihood to be exported in the future.  Our results show that relatedness is scale-dependent:  the best assessments are obtained by using machine learning on the same typology of data one wants to predict. Moreover, we found that while relatedness measures based on country data are not suitable for firms, firm-level data are very informative also for the development of countries. In this sense, models built on firm data provide a better assessment of relatedness. We also discuss the effect of using parameter optimization and community detection algorithms to identify clusters of related companies and products, finding that a partition into a higher number of blocks decreases the computational time while maintaining a prediction performance well above the network-based benchmarks.\end{abstract}
		
		
	\end{frontmatter}
	
	\section{Introduction}
	Relatedness \cite{hidalgo2018principle}, a key tool of the Economic Complexity framework \cite{hidalgo2021economic}, refers either to the similarity between two economic activities or between an activity and an economic actor. As such, it is also known as coherence \cite{teece1994understanding} in the standard economic literature. This concept can be easily applied to different sets of such activities, such as the export basket of countries \cite{hidalgo2007product,zaccaria2014taxonomy}, the technology portfolios of companies\cite{breschi2003knowledge,nesta2005coherence,pugliese2019coherent}, or regional diversification patterns \cite{neffke2011regions}. In these cases, relatedness is a measure of the feasibility of an activity (e.g., exporting a product) with respect to what an economic actor already does. This tool is, at present, widely adopted by policymakers and institutions such as the World Bank Group \cite{lin2020african,zaccaria2018integrating} and the European Commission \cite{pugliese2020economic,pugliese2021economic}
	to inform governments and the private sector with respect to industrial and innovation policy, both at a country and regional level.\\
	Being relatedness a general concept, the precise way to assess the similarity between two activities or the feasibility of an activity for an economic actor is, a priori, not determined. As a consequence, various formulations co-exist in the literature; most of them are however related to the so-called co-occurrences, that is counting how many countries are exporting a couple of products (the more the counting, the more related the two products will be). This is equivalent to projecting the input data (a bipartite economic actor-activity network, typically a country-product network) into one of the two layers \cite{zhou2007bipartite}, usually the economic activities, for instance the exported products. For example, the projection of the bipartite country-product network into the layer of products gives rise to a monopartite network of products. Among the various possibilities, Teece et al. \cite{teece1994understanding} proposed to use the t-statistics of co-occurrences in industries with respect to a randomized diversification of firms. Hidalgo et al. \cite{hidalgo2007product} introduced the Product Space, in which the co-occurrences of exported products are normalized with respect to their ubiquity. Zaccaria et al. \cite{zaccaria2014taxonomy} normalized the co-occurrences with respect to both ubiquity and diversification, to take into account the nested structure of the country-product network, the idea being that highly diversified countries carry a relatively small amount of information. In the cases above, the result is an almost fully connected network, whose pictorial representation is not very informative and, most importantly, no effort is present to remove any possible noise. So various attempts to filter such projections are present in the literature. Saracco et al. \cite{saracco2017inferring} proposed to statistically validate the single links with respect to a null configuration model \cite{squartini2011analytical}. Cimini et al. \cite{cimini2021meta} however showed that the adoption of different null models leads to different filtered networks; see also Dosi et al. \cite{dosi2017firms} and Bottazzi and Pirino \cite{bottazzi2010measuring} for a critical discussion of the consequences of using different null models while computing relatedness.\\
	This situation calls for a framework to systematically compare and validate the different relatedness measures. Our proposal is to use an out-of-sample prediction task for this purpose. Tacchella et al. \cite{tacchella2021relatedness} and Straccamore et al. \cite{straccamore2021will} have shown that standard co-occurrences methods perform worse than auto-correlation benchmarks, and that tree-based machine learning algorithms such as Random Forest \cite{breiman1996bagging,breiman2001random} provide the present state-of-the-art with respect to the assessment of relatedness. Albora et al. \cite{albora2021product} described this approach in detail, providing a comparison between different machine learning algorithms.\\
	Having established that the relatedness between a country and a product is better assessed by means of machine learning, a natural question arises, namely whether the country data provides an optimal assessment, given the fact that companies, and not countries, are actually
	producing the exported products. Moreover, often recommendations are given to the private sector, so one could expect that algorithms should be trained on companies, and not countries. In this paper, we provide a systematic and quantitative comparison of machine learning and network-based approaches to forecasting new products both at country and firm-level. In particular, we will leverage a database of more than 70000 Italian firms and compare it with country-level data, providing a cross-database analysis for both training and testing. Our results provide quantitative evidence about which algorithm and which database should be used to optimally assess relatedness; moreover, we are able to economically motivate these results by investigating the different structure of the two databases and how the algorithms extract the relevant information.
	
	\section{Materials and Methods}
	In this section we discuss our database, the metrics to compute relatedness, and the testing procedure.
	\subsection{Firm-level Data}
	The Italian National Institute of Statistics (www.istat.it) provided data about the export of all Italian firms. After a preliminary cleaning procedure, we have 71826 Italian firms exporting at least two products in the period between 1996 and 2017 and at least one product in every year from the one on which they exported their first product till 2017. The exported products are classified according to the UN-COMTRADE (comtrade.un.org) Harmonized System, 1992 edition. This is a hierarchical classification, encoded by a number of digits corresponding to different levels of aggregation. For our investigation we use 4 digits, corresponding to 1233 codes, defining as many different products. This data can be organized as a set of temporal bipartite networks, one for each year, linking Italian firms with their exported products; at first, the weight of each link is the export volume.
	This is equivalent to defining 22 matrices $E(y)$ (y=1996...2017) of size $71826\times1233$, where each row represents a firm and each column a product. The element $E_{fp}(y)$ is the volume of product $p$ (expressed in euros) that firm $f$ exported during the year $y$.\\
	\subsection{Country-level Data}
	country-level data comes from UN-COMTRADE database (comtrade.un.org) and consist in the exports of 169 countries in the period between 1996 and 2017. All the considerations made above still apply. In order to match this data with the firm-level, we use only the 1233 products that are present also in the Italian firms' data. We point out that, being Italian economy highly diversified, this corresponds to discarding less than 1\% of products (0.8\%). So at country-level we have 22 matrices (one per year) of size $169\times1233$.
	
	\subsection{Data Preprocessing}
	Since export volumes strongly depend on the size of both the economic actor (country or firm) and the specific product, the direct use of this quantity would introduce a strong bias.
	The usual solution in the economic complexity literature \cite{hidalgo2007product,Tacchella}
	is to compute the RCA values (Revealed Comparative Advantage) introduced by Balassa \cite{balassa1965trade} defined as:
	\begin{equation}
		RCA_{fp}(y)=\frac{E_{fp}(y)}{\sum_{p'}E_{fp'}(y)}\frac{\sum_{f'}\sum_{p'}E_{f'p'}(y)}{\sum_{f'}E_{f'p}(y)}
	\end{equation}
	In this way the export are normalized with respect to both the total export of the firm and the product and, using a physics jargon, we go from the extensive variable $E$ to an intensive one. In order to have a binary variable, we say that a product $p$ is (competitively) exported by a firm $f$ if its RCA is greater than 1 and with this threshold we define the binary matrix M
	\begin{equation}
		M_{fp}(y) =
		\begin{cases}
			1 ~~~\text{ if } RCA_{fp}(y)\geq1\\[3mm]
			0 ~~~\text{ if } RCA_{fp}(y)<1
		\end{cases}	
	\end{equation}

	\subsection{Relatedness Measures}
	The first aim of our analysis is to compare different approaches to measure the relatedness between firms and products, that is how much a firm is close to being able to export a product. This is something largely studied when the economic actors are not firms but countries; as discussed in the introduction, two types of approaches exist: complex networks \cite{hidalgo2007product,teece1994understanding,zaccaria2014taxonomy} and supervised machine learning algorithms \cite{tacchella2021relatedness,albora2021product,straccamore2021will,che2020intelligent}. Here we compare these methods in an out-of-sample forecast exercise both at country and firm-level, in which we assume that exporting more related products is easier. The output of both the network-based and the machine learning approach is an S matrix in which the element $S_{fp}$ is the relatedness between firm $f$ and product $p$.\\
	\subsubsection{Networks Models}
	Traditionally, in order to measure the relatedness between a country and a product, one starts with a measure of the similarity or proximity between products, that can be visualized as a network of products. The next step is the computation of the density \cite{hidalgo2007product} or coherence \cite{pugliese2019coherent}: the average similarity between the target product and the ones already exported by the target country. This is what we will call the relatedness between a country and a product.\\
	To compute the proximity between two products one counts how many countries export both, that is, the number of co-occurrences. The weight of the link of the resulting network of products is this quantity possibly divided by a normalization factor. According to the latter, we can define different types of networks. In the product space \cite{hidalgo2007product} we divide the number of co-occurrences with the maximum ubiquity between the two products (i.e. how many countries export that product). In formula:
	\begin{equation}
		B^{PS}_{pp'}=\frac{1}{\max(u_p,u_{p'})}\sum_{c}M_{cp}M_{cp'}
	\end{equation} 
	where $u_p=\sum_{c}M_{cp}$. However, the co-occurrence of products in a country that exports almost all the products is not so relevant like the one in a country that exports few products. This is a relevant problem, given the nested structure of the matrix $M$ \cite{mariani2019nestedness}. An improvement that takes into account this factor is the taxonomy network \cite{zaccaria2014taxonomy} in which each co-occurrence is also normalized with respect to the diversification of the country $d_c=\sum_pM_{cp}$:
	\begin{equation}
		B^{TN}_{pp'}=\frac{1}{\max(u_p,u_{p'})}\sum_{c}\frac{M_{cp}M_{cp'}}{d_c}.
	\end{equation}
	Once we have the network B we define the relatedness $S$ between a country and a product by using the density \cite{hidalgo2007product}:
	\begin{equation}
		S_{cp}=\frac{\sum_{p'}M_{cp'}B_{pp'}}{\sum_{p'}B_{pp'}}.
	\end{equation}
	In practice, we sum the export matrices $E$ from 1996 to 2012 to obtain the total export either of the countries or of the firms, we compute the RCA and M values from the resulting matrix and we estimate all the weights of the links B. Then we apply the formula above using the M matrix of 2012 to compute the relatedness S either for the countries or for the firms that belong to the test set defined below.
	
	\subsubsection{Random Forest}
	The measure of relatedness given by the use of machine learning algorithms based on decision trees has been shown to provide a better assessment of the probability for the future exports of countries than network-based approaches \cite{tacchella2021relatedness}. In particular, it has been shown that random forest \cite{breiman2001random} and XGBoost  \cite{chen2016xgboost} are the most performing algorithms for the task of assessing relatedness \cite{albora2021product}. In this paper, we decided to adopt the random forest (RF) since, even if with country-level data XGBoost gets slightly superior results \cite{albora2021product}, the computational time required to train a random forest is much lower, so it allows us to make a more complete analysis with a tuning of the hyperparameters and, as we will see, the use of community detection algorithms.\\
	Before talking about the training procedure we need to talk about how we split the data. When we are working with country-level data we use all the countries both for the training and the testing of the algorithms, but when we work with firm-level data we split the firms into three datasets: 20000 firms are used to train the algorithm, another 20000 are used for the validation procedure to make the tuning of the hyperparameters and the remaining 31826 firms are used to do the test. The reason why we do not use all the firms in the training like we do with country-level data is that firms are much more than countries and using all of them would increase the computational time. The firms in each of the three datasets are chosen randomly.\\
	For each product $p$ we build a random forest that has the task to predict if firms (or countries) will export $p$ after 5 years.  In the case of firm-level data, during the training procedure the features are given by the concatenation of the 12 RCA matrices $20000\times1233$ between 1996 and 2007 in order to have a single matrix $240000\times1233$ in which each row contains the RCA values in a year $y$ of a firm that belongs to the training set. The labels of the training are given by the concatenation of the column $p$ of the M matrices from 2001 to 2012. So, during the training, the model learns if with a certain configuration of the RCA values in year $y$ (i.e. the export basket of a firm) the firm can start to export the product $p$ after 5 years.\\
	There are some hyperparameters that can be optimized to improve the performance of the random forest and to avoid overfitting. Here we take into consideration \textit{max depth} and \textit{min samples leaf} \cite{geron1991handson}. The value of the first hyperparameter regulates the maximum depth of the tree, if one of the trees of the random forest reaches this value during its construction, its training stops even if not all the training samples are perfectly classified. So if the trees are very deep and the too many splits bring to overfitting, to avoid it we can lower the \textit{max depth}. The value of the second hyperparameter regulates the minimum value of training samples that a leaf node must have. If during the training the algorithm finds a split that creates a node with less samples than the value of \textit{min sample leaf}, this split is discarded. So an high value of \textit{min sample leaf} prevents the random forest from creating nodes with few training samples that are the ones that bring to overfitting. In order to optimize the hyperparameters we train different random forests varying the value of one of the hyperparameters at once, then we perform a test using the firms of a validation set and we choose the value of the hyperparameter that brings to the higher best F1 score.\\
	Once found the optimal values of \textit{max depth} and \textit{min samples leaf}, we train a random forest with these values and we do the test with the firms in the test set. Giving the RCA values of 2012 as input to the random forest related to a product $p$, it returns a vector with the column $p$ of the S matrix, so with all the 1233 random forests we build the whole S matrix for the firms needed in the test set. The same procedure can be repeated for the 169 countries.\\
	
	In the results section we are going to compare the models trained on countries and the ones trained on firms to make predictions about countries or firms. This cross-test may change 
	depending on the peculiar value used as an input. In particular, one may use directly RCA instead of the binary M. Our idea is to always use the most informative input. When we train the random forest on firm-level data and we also do the test on firm-level data we use RCA, however if we train the model on firm-level data to make predictions on countries, we use M values as input variables. The reason is that countries and firms have very different RCA values since they are very different objects. Indeed, the average nonzero RCA value for firms is about 500, while for countries it is 2, so for a firm RCA = 4 is a low value, while for a country it is well above average. So in what follows, when we show the results of a model that has been trained on firms to make predictions on countries the input variable is the M values, instead, if we make predictions on firms the input variable is RCA. The same goes for the models trained on country-level data, if the predictions are on firms we use M, while if they are on countries we use RCA.\\
	\subsection{Testing procedure}
	In order to test the goodness of the relatedness assessment we assume that firms (or countries) will export in the future products with the higher S (relatedness) values. In particular, we build the models discussed above by using data from 1996 to 2012, from which we compute the S matrix. The comparison of these relatedness measures with the M(2017) matrix can be seen as evaluating the output of a binary classifier. In particular, the hypothesis is that the higher $S_{fp}$ the more likely will be that firm $f$ will start to export product $p$. This is analogous to common machine learning classification exercises \cite{geron1991handson,kotsiantis2007supervised}, so in order to compare the goodness of the different relatedness metrics, we can use the performance indicators we introduce in section 2.6. However, given the strong self-correlation of the export matrices, what interests us is not predicting if firms will export products they already export, but if the will export new products. For this reason, when we do the comparison of the S matrix with M(2017), we consider only the activations of new products, or in other words we remove the elements (f,p) that do not satisfy this requirement:
	\begin{equation}
		RCA_{fp}(2012) < 0.25
	\end{equation}
	In this way we look at how good the model is in predicting the activation of new products by firms. The value 0.25 of the threshold follows \cite{tacchella2021relatedness,albora2021product}. The idea is that using a threshold $RCA<1$ would increase the noise in the test set given by products whose RCA value for a firm fluctuates around 1. Moreover, predicting that a firm can be competitive in the export of a product on which its RCA value is already close to 1 is less interesting with respect to a firm genuinely becoming competitive. In order to check the robustness of our findings, we repeated the forecast exercise using different values of the threshold finding similar results. As already said, to build the model we use a set of 20000 firms (training set) and to make the comparison with M(2017) we use a separate set of 31826 firms (test set), in this way the test is out of sample because during the construction of the model no year between 2013 and 2017 is used and the firms from which the model learns are not the ones on which we make predictions.\\
	When we work with countries there are some differences, since the countries are only 169 we use all of them both to train the model and to do the test. The other difference is that with country-level data we use a stronger definition of activation in order to align the results with the ones published in \cite{albora2021product}:
	\begin{equation}
		RCA_{cp}(y) < 0.25~~~\forall y\in[1996,2012]
	\end{equation}
	The main reason we chose to use two different definitions is that firms can be absent in toto until a certain year and then start exporting, while countries always export at least one product in the years from 1996 to 2012. 
	\subsection{Performance indicators}
	In this section we describe the indicators which quantify the goodness of the forecast.
	When evaluating a binary classification the choice of the performance indicator depends on both the research purpose and the database structure \cite{powers2011evaluation,caruana2006empirical}. In our case, the fraction of ones in the M matrices of firms is only the 0.4\% of the total elements, the remaining being equal to zero. So we have to deal with a very high class imbalance and for this reason we have to carefully choose our performance indicators. For instance, if we would use indicators that involve the true negatives like accuracy, they would get very high values because even if we do not guess any True Positive, the number of True Negatives will likely be huge.
	Here we make a quick description of the indicators we use:
	\begin{itemize}
		\item \textbf{Precision} \cite{powers2011evaluation} It is the number of True Positives divided by all the Positives, respectively how many products we guess to be competitively exported by firms after 5 years and how many products we expected to be exported considering also the wrong predictions.
		\item \textbf{P@K} The precision@K corresponds to the fraction of the top K positives that are correctly predicted or, in other words, the fraction of elements that the model guesses if we ask it to tell us the K most probable positives.
		\item \textbf{mP@K} The mean Precision@K is computed considering only the first K predicted products separately for each firm, then we look at how much of them are correct and finally we average on the firms. By using mP@K we quantify the correctness of our possible recommendations of K products, on average, for a firm. In this sense mP@K is a local measure of performance, while P@K takes into account the whole matrix.
		\item \textbf{Recall} \cite{powers2011evaluation} It is the number of True Positives divided by the sum of True Positives and False Negatives.
		\item \textbf{F1 Score} \cite{dice1945measures,van1974foundation} Precision and Recall vary by changing the scores' binarization threshold (the value we use to define Positives and Negatives). Usually the higher it is the Precision, the lower it is the Recall and vice versa. F1 score is a harmonic mean of these two metrics and its value is high only if both of them are relatively high. We define as \textbf{best F1} score the F1 score computed by finding the threshold that maximizes it.
		\item \textbf{ROC-AUC} \cite{hanley1982meaning, hajian2013receiver} It is computed ranking all the scores and computing for each possible threshold the True Positive Rate (TPR) and the False Positive Rate (FPR). In this way we draw a curve in the TPR/FPR plane and the ROC-AUC corresponds to the area under this curve. It can be seen as the probability that, if we randomly select a positive and a negative element, the first will receive a higher score \cite{fawcett2006introduction}. With highly imbalanced data, due to the high number of True Negatives, the ROC-AUC tends to give too optimistic results  \cite{saito2015precision, fernandez2018learning}.
		For a random classifier, ROC-AUC=0.5.
		\item \textbf{AUC-PR} The area under the precision-recall curve is the area under the curve that, in the plane defined by Precision and Recall, is obtained by varying the scores' binarization threshold. Since True Negatives are not considered, its value is not misled by the class imbalance \cite{saito2015precision}.
		\item \textbf{MCC} \cite{matthews1975comparison} Matthews' correlation is computed using the scores' binarization threshold that maximizes the F1 score. It is a metric that takes into account all the four classes of the confusion matrix and also the class imbalance issue \cite{chicco2020advantages,boughorbel2017optimal}.
	\end{itemize}
	Precison, recall, F1 score and MCC require a threshold to define if the value of the score should be associated with a positive or negative prediction. In this cases, we chose the threshold that maximizes the F1 score.
	\section{Results}
	\subsection{Random Forest on Firms vs Product Space on Countries: worked examples}
	\label{res1}
\begin{figure*}
		\centering
		\includegraphics[width=0.49\textwidth]{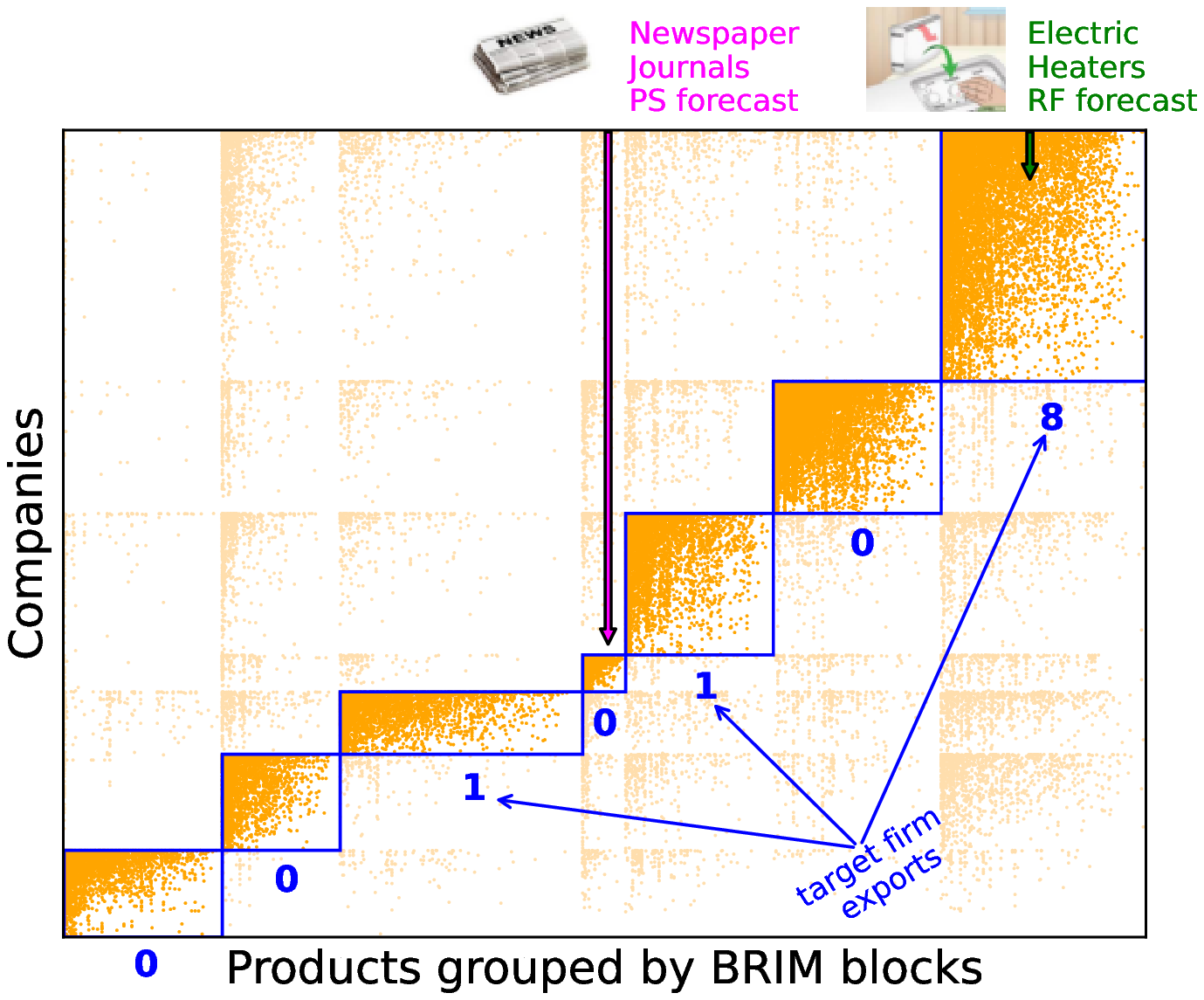}
		\includegraphics[width=0.49\textwidth]{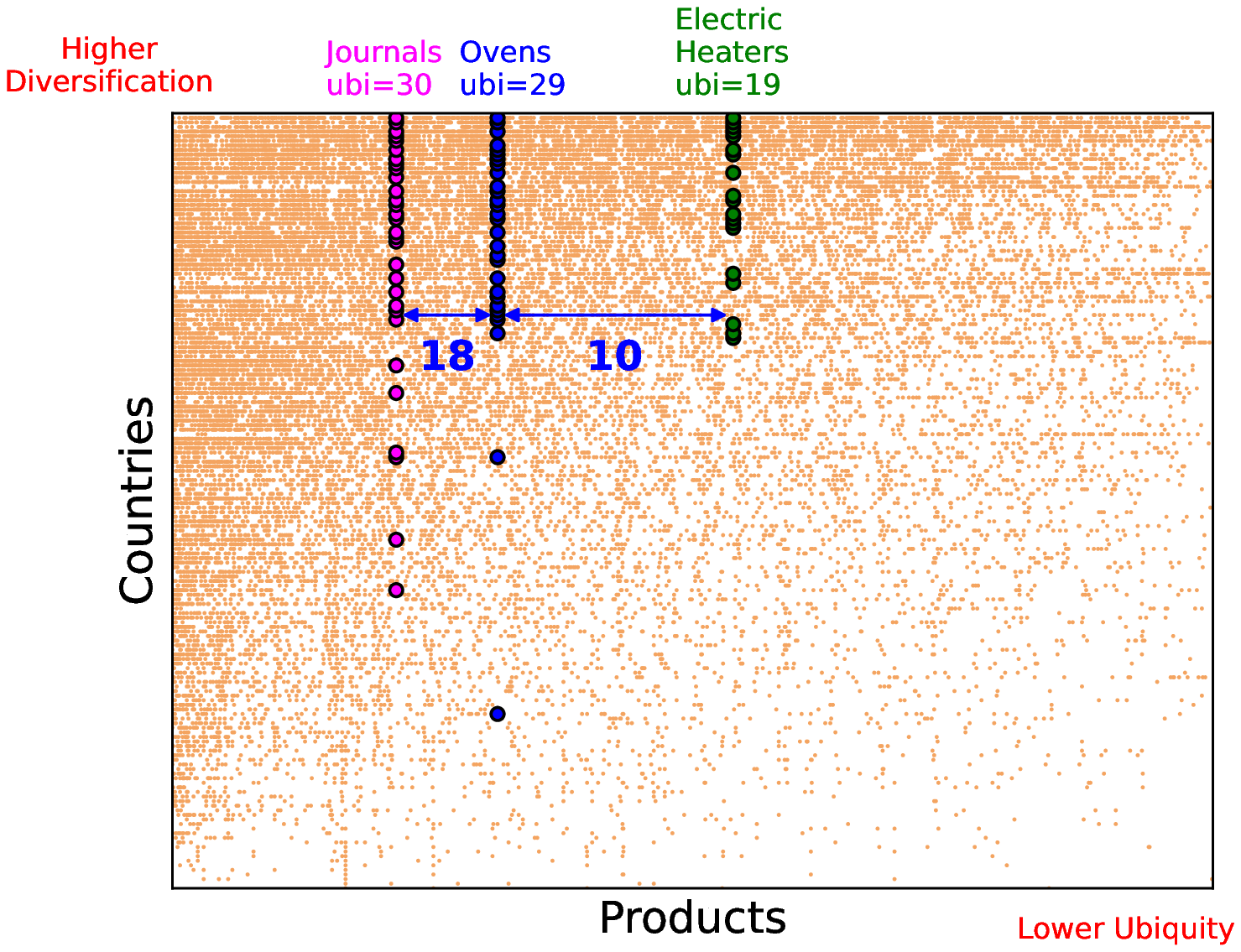}
		\caption{Visual representation of the predictions given by product space built on countries and random forest built on firms. Left plot: matrix representation of the firm-product network. The products are grouped in blocks detected by the BRIM community detection algorithm. Under each block we report the number of products that belong to the exports of the target firm, firm B (selling large kitchens). The random forest prediction (Electric Heaters, green arrow) falls in the block in which the target firm has 8 products; the magenta arrow points to the most probable future export (Newspapers) according to a product space model built on country-level data and falls in a block in which the firm has no products. Right plot: country-product network. The products are sorted in decreasing order of ubiquity while the countries are sorted in increasing order of diversification. We highlighted in blue a product exported by the target firm, Ovens. As expected, the product space forecast for the target firm is a more ubiquitous (simpler) product that is exported by many countries and so has many co-occurrences (18) with the ovens. The random forest forecast is a less ubiquitous (more complex) product and for this reason it has fewer co-occurrences (10) with the ovens}
		\label{fig: giornali}
	\end{figure*}

	In this section we present data-driven examples to	compare the different forecasts on future exports of the same Italian firms given by i) a product space (PS) approach built on country-level data and ii) a random forest (RF) trained on firm-level data. What we want to highlight is not only that the RF has more predictive power than the PS, as we will quantify better later, but also that the choice of the data with which we build the model is fundamental.\\
	The first reason why a model that is trained on country-level data produces worse forecasts on firms' future exports is that what is similar from the point of view of a country is oftentimes not similar from the point of view of a firm too. To clarify this point we take an example from the data: in 2012 a historic firm (firm A) that deals with jewelry, in particular with corals, exported the products in table \ref{tab: gioielli}.
	\begin{table*}
		\centering
		\begin{tabular}{c|c|c|}
			code&description&HS section\\
			\hline
			\hline
			0508&Coral and similar materials&1\\
			\hline
			7103&Precious (excluding diamond) and semi-precious stone&14\\
			\hline
			7117&Imitation jewelry&14\\
			\hline
			9601&Ivory, tortoise-shell, horn, coral and other animal carving material and articles&20\\
			\hline
		\end{tabular}
		\caption{Example of exported products by firm A. Clearly, this firm deals with jewelry and corals and in the last 20 years it exported these products for about 10 million euros .}
		\label{tab: gioielli}
	\end{table*}
	The RF trained on firm-level data correctly predicts that this firm in 2017 will export the product with HS code 7113 that is \textit{Jewellery articles of precious metal or of metal clad with precious metal}. However, PS wrongly predicts the product 0307 that is \textit{Molluscs, whether in shell or not, live, fresh, chilled, frozen}. So while the RF understood that firm A deals with jewelry, the PS recommended molluscs. The reason is that the firm exports corals and in a country where there are corals there are also molluscs. If you are a country and you have the sea you will have both firms that treat corals to export jewelry and firms that treat and export molluscs and other seafood products, but if you are a single firm either you treat jewelry or you treat seafood products. So the PS found a relation between corals and molluscs that is relevant only if you are a country.\\
	Another reason why the PS built on country data is not suitable for forecasting future exports of firms lies in the specialized nature of firms. A big difference between a country and a firm is that the first tends to diversify and export as many products as possible, while the second is specialized in a category of products \cite{bruno2018colombian}. When one builds the PS using country-level data one observes and counts the co-occurrences between different products following the nested pattern of the export data (see Figure \ref{fig: giornali}, left). In particular, simple products are exported by almost all countries and so the PS counts a large number of co-occurrences with the complex products. On the contrary, complex products are exported by only those countries that export most of the products, including the less sophisticated ones. So one can expect that, by using such a model, when one tries to forecast the future exports of a firm specialized in complex products one can wrongly predict that firm to export random simple products.\\
	In Figure \ref{fig: giornali} we show a real example to clarify this point.
	On the left plot we show the adjacency matrix of the bipartite firm-product network. Companies are on the vertical axis while products are on the horizontal axis. The ordering of rows and columns is given by the BRIM community detection algorithm \cite{barber2007modularity}. The orange points highlight the products that firms exported in 2012. The evident modular structure of the firm-product matrix reflects the specialized nature of firms. We selected as a target firm an important company specialized in the design and production of kitchens (firm B) that in the last 20 years exported products for more than 500 million euros. The firm exported ten products in 2012, eight of which belong to the same block. According to the PS model built on country-level data, the most related product, and so its best guess for a future export of the target firm, is \textit{newspaper, journals and periodicals} (magenta arrow). The un-relatedness of this product with respect to the target firm is self-evident. Moreover, newspapers belong to a different block than the one on which the firm is active, in particular, no products of this block are exported by the firm, and obviously, if we check these predictions, in 2017 the firm  will not export journals. When we ask the same question to a RF built on firm-level data, the output is \textit{electric water, space, soils heaters}, which belongs to the same block where the target firm already exports 8 products. Moreover, in 2017 the firm will start exporting also electric heaters. It is evident that, while the RF understood what category of products the firm deals with, the PS did not. The fact is that, being built on country-level data, the PS did not learn the specialized nature of firms; what it knows is that among the countries there are many co-occurrences between newspapers and the products of the target firm, but this is only a consequence of the fact that many countries export newspapers since it is a simple product. We can see this from the right plot. On the horizontal axis we report the products in decreasing order of ubiquity (that is highly anti-correlated with the complexity) and on the vertical axis we report the countries in increasing order of diversification. Newspapers in 2012 are exported by 30 countries; they are less sophisticated than electric heaters, which are exported by 18 countries. We highlighted with blue points one of the ten products exported in 2012 by the target firm, that is \textit{Furnaces and ovens; industrial or laboratory} and it is exported by 29 countries. Journals and ovens have 18 co-occurrences, while electric heaters and ovens have only 10 co-occurrences, so it is evident that PS built on these data learns that ovens are more similar to journals than to electric heaters because of the nested structure of the matrix.\\
	
	\subsection{Firm-level relatedness outperforms}
\begin{figure*}
		\centering
		\includegraphics[width=0.495\textwidth]{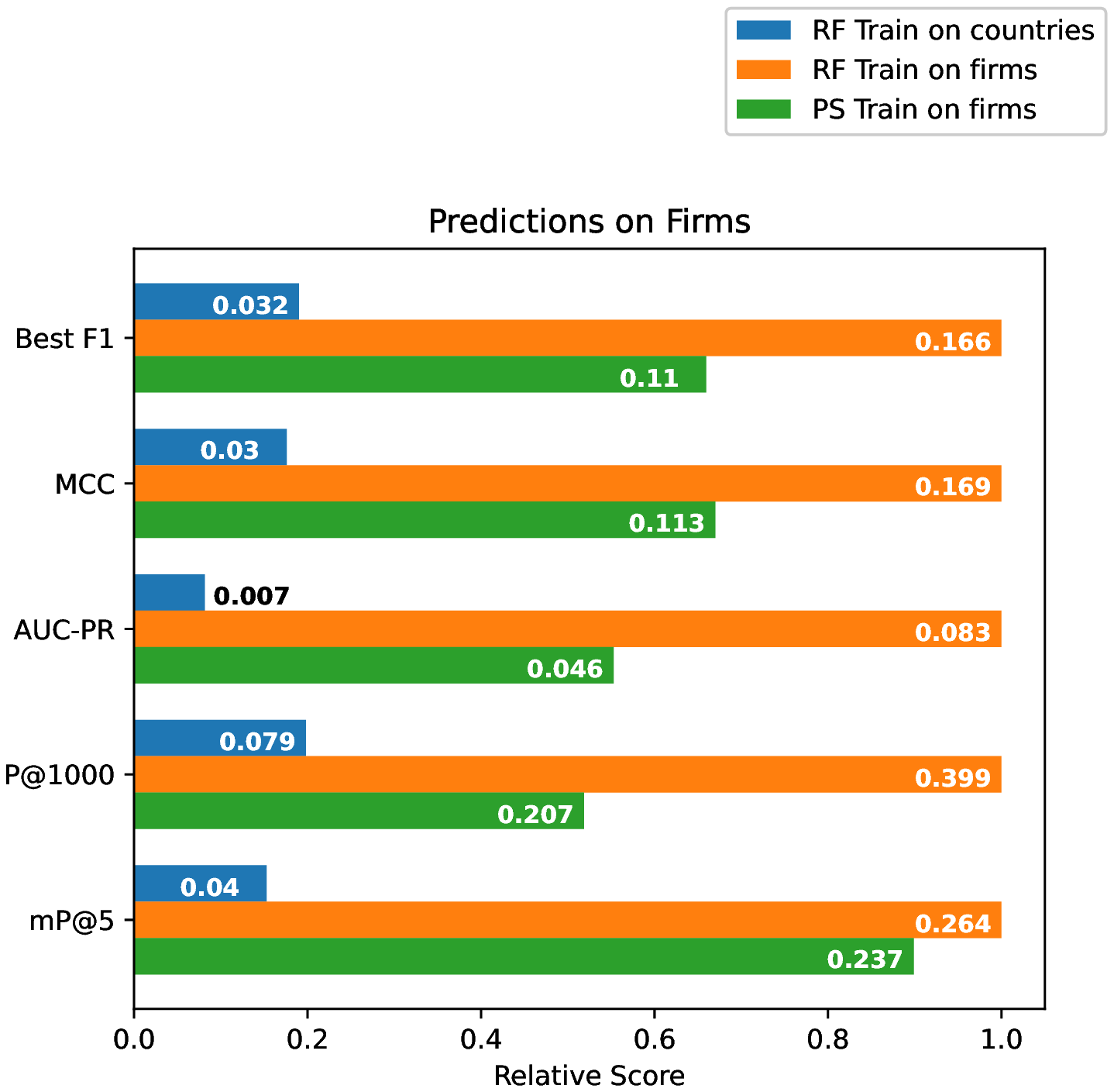}
		\includegraphics[width=0.495\textwidth]{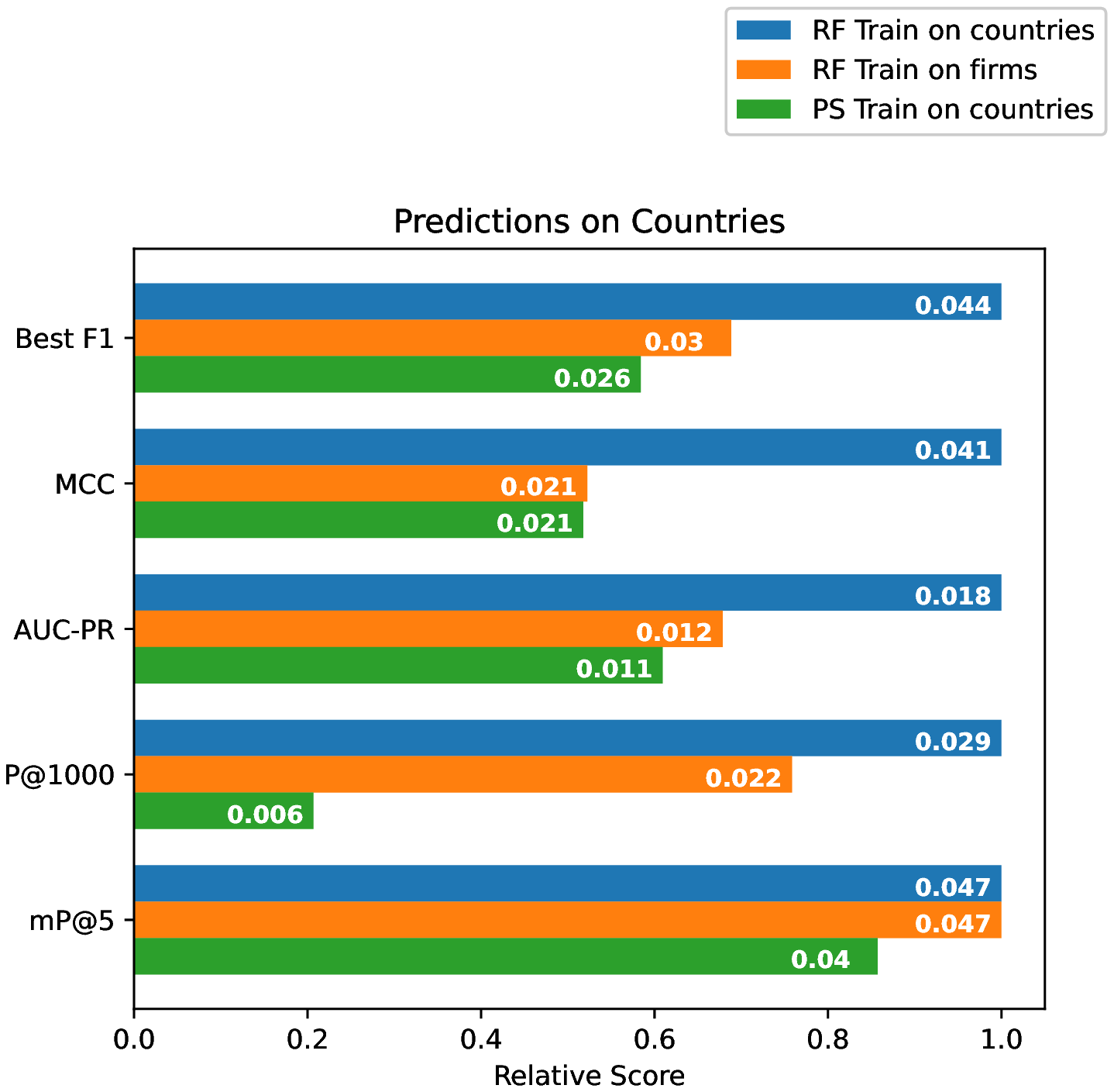}
		\caption{Comparison using different metrics between random forest built on firms and built on countries when we predict future exports of both firms (left) and countries (right). The orange bars refers to a random forest that has been trained on firm-level data, the blue bar refers to a random forest trained on country-level data and the green bar refers to a product space built on firms (left) or country (right) data. The best assessment of relatedness at country and at firm-level is given by the relative random forest. However, the relatedness computed at firm-level performs better than the product space also at country-level.}
		\label{fig: comparison}
	\end{figure*}

	In the previous section we show that the modular structure of the firms' database allows for a better quantification of relatedness with respect to the nested structure of the countries database.
	Now we compare a RF trained on country-level data with a RF built on firm-level data and we compare the performance of the two models on the prediction of both the future exports of firms and the future exports of countries. The assumption is that a higher relatedness implies, on average, a higher probability that the country or firm will export the target product. While in the previous section we presented specific examples, here we provide a general, quantitative evaluation of the model's predictions, always using PS as a benchmark.\\	
	In figure \ref{fig: comparison} on the left we show the performance of different models when we try to predict the future exports of firms. The blue bars refer to a RF trained on country-level data,  the orange ones refer to a RF trained on firm-level data and the green ones refer to a PS built on firm-level data. We show different indicators to show the robustness of our results; we rescaled their values to allow a visual comparison. As expected from the qualitative discussion of the previous section, RF trained on firm-level data is by far the best choice; in particular, the model trained on country-level data is totally unsuitable to make predictions about the future exports of firms. On the right plot, the performance refers to a prediction of the future exports of countries and, accordingly, now the PS is built on country-level data. This time the RF trained on country-level performs better and the reason is that there are relationships between products that are not present at firm-level, like the ones discussed above: a model built on country-level data is trained using observations like marine countries that export both corals and molluscs, while a model built on firm-level data is trained using samples that, if exporting corals, will never export molluscs. So the former will use the co-occurrences between molluscs and corals while the latter will not learn them, because at firm-level this relationship does not exist. However, the difference between country-level and firm-level RF is smaller with respect to the plot on the left. Strikingly, the RF trained on firm-level data performs better than the PS trained on country-level data even if we are making predictions on countries. We can conclude that machine learning models trained at firm-level are able to extract a measure of relatedness that is relevant also at country-level, and in particular, more relevant than the information that the PS is able to extract even at country-level.
	In conclusion, firm-level data provide a relatedness measure that is objectively better than the one given by country-level data.
	
	\subsection{Model comparison}
\begin{figure*}
		\centering
		\includegraphics[width=0.8\textwidth]{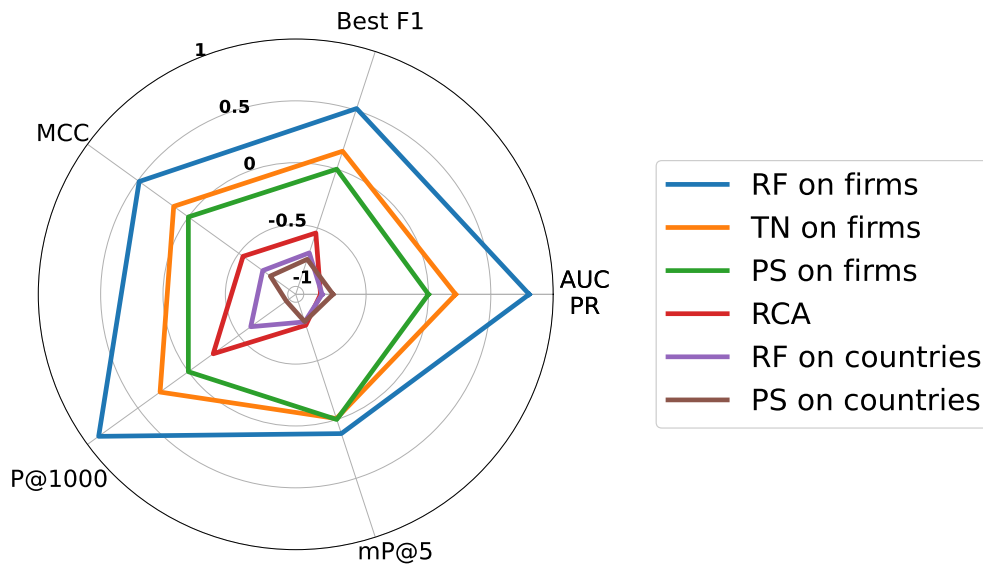}
		\caption{Model comparison in assessing firms' relatedness. Each vertex refers to a performance indicator normalized with respect to a product space built on firms. Models built on countries perform worse than the RCA benchmark. Taxonomy network performs better than the product space. The random forest trained on firms results to be the best model overall.}
		\label{fig: radar}
	\end{figure*}

	In this section we compare different models to predict the exports of firms. In particular, we compare RF with network models like PS and taxonomy network (TN) and we will also show the results obtained by using a quasi-trivial benchmark, RCA itself. Indeed, one may think to assume an auto-correlation model and consider as prediction score the RCA value in year $y=2012$ of a firm $f$ on a product $p$. The resulting matrix S is the relatedness between firms and products and it is treated exactly in the way we show in sections 2.5 and 2.6. The higher the RCA, the higher the likelihood that $f$ will start to export $p$. \\
	In figure \ref{fig: radar} we show a radar plot in which the performance of the PS (green line) built on firms is used to normalize the other scores.
	Each of the vertices in the radar plot refers to a different metric and the area of each polygon is a proxy of the total performance of the corresponding model. The brown and purple lines refer respectively to the PS and the RF built on country-level data. They not only perform worse than all models trained on firm-level data, but they underperform the RCA predictions too (red line). The orange line is the TN built on firm-level data, which is slightly better than the PS. The RF built on firms vastly outperforms all the other models.\\
	In table \ref{tab: results} we compare the results of the models using all the performance metrics described in the method section. All the models are trained on firm-level data. The large majority of the classified elements are True Negatives because of the high class imbalance of the problem; so metrics that involve the True Negatives like Accuracy should be avoided since they would give very high results only because it is very easy to predict a True Negative. For instance, the ROC-AUC is very high for all the models except the RCA one and provides misleading results. 

	\begin{table*}
		\centering
		\begin{tabular}{|c|c|c|c|c|}
			\hline
			&RCA&Random Forest&Product Space&Taxonomy Network\\
			\hline
			Best F1&0.050&0.166&0.110&0.126\\
			\hline
			AUC-PR&0.006&0.083&0.046&0.056\\
			\hline
			ROC-AUC&0.519&0.936&0.919&0.928\\
			\hline
			Precision@1000&0.156&0.399&0.207&0.265\\
			\hline
			mP@5&0.047&0.264&0.237&0.235\\
			\hline
			MCC&0.052&0.169&0.113&0.130\\
			\hline
			TP&2703&15584&11482&12706\\
			\hline
			FP&33387&100292&126310&116773\\
			\hline
			FN&69037&56156&60258&59034\\
			\hline
			TN&38952640&38885735&38859717&38869254\\
			\hline
		\end{tabular}
		\caption{Prediction performance of firm-based models.
			Our problem is characterized by a huge number of True Negatives, and this precludes the use of metrics such as accuracy and ROC-AUC. In any case, the random forest outperforms all other models.}
		\label{tab: results}
	\end{table*}
	
	\subsection{Random forest optimization: leveraging modular structure and hyperparameters}
\begin{figure*}
		\centering
		\includegraphics[width=1\textwidth]{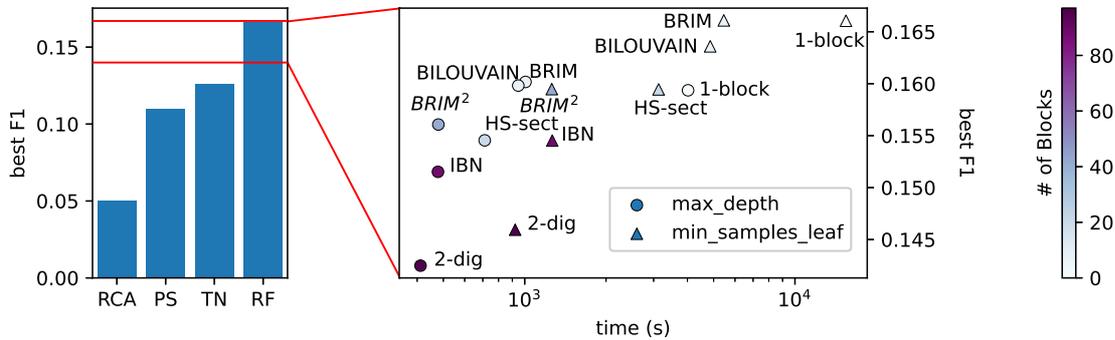}
		\caption{Performance of the random forest by varying the hyperparameters and the partitions that define the feature vector given as an input for each product. The results are optimized by tuning of \textit{max depth} (circles) or \textit{min samples leaf} (triangles). We also write the adopted partition and the color of the points represent the resulting number of blocks. On the horizontal axis we report training time and on the vertical axis a performance indicator. Using more data (larger blocks for each product) provides better performance but takes longer time for training; smaller blocks lead to faster but less precise results. In any case, as evinced by the zoom given by the red lines on the plot on the left, random forest always outperforms the other models.}
		\label{fig: blocks}
	\end{figure*}

	As we have shown in the previous sections, the firm-product export matrix has a modular structure.
	In this section we investigate if such structure can be exploited to improve the prediction performance of machine learning; in particular, we perform a community detection on the bipartite graph and we train each RF by giving as input only the RCA values of the products that belong to the same block of the target product. Community detection is characterized by a number of different algorithms \cite{fortunato2010community}, so it is natural to consider various possibilities to build the partitions. Two natural block decompositions can be derived from the hierarchical structure of the Harmonized System classification: the 1233 4-digits products can be organized in 21 sections or 96 chapters (the latter corresponding to a 2-digit aggregation level). Moreover, using community detection algorithms we can find other partitions; in this paper we use: BRIM \cite{barber2007modularity}, BILOUVAIN \cite{traag2011narrow} and IBN \cite{sole2018revealing}.\\
	To which extent the performance of RF can be improved by tuning its hyperparameters is still debated \cite{probst2019hyperparameters}. Here we discuss the effect of changing two of these parameters in assessing relatedness: \textit{max depth} and \textit{min samples leaf}. The default values are \textit{max depth} = $\infty$ and \textit{min samples leaf} = 1 \cite{geron1991handson}. However, this choice can lead to overfitting, because each decision tree is expanded up to a perfect classification of the training sample.\\
	In figure \ref{fig: blocks} we compare the performance of different RF models trained using the partitions given by different community detection models and with different choices of the two hyperparameters. In particular, we compare both the prediction performance (quantified by the best F1 score) and computational time on a standard desktop computer.
	
	On the right plot each point represents one RF model with a \textit{max depth} (circles) or \textit{min samples leaf} (triangles) optimization. Close to each point we report the partition criteria that defines the blocks seen by the RFs: since we train one model for each product, each product is predicted by using the block it belongs to. Here 1-block means that we use all available data (so all products see all products) and $BRIM^2$ means that we applied the BRIM algorithm twice. On the horizontal axis we report the training time and on the vertical axis the best F1 score. The color of the points represent the number of blocks resulting by the corresponding partition. On the left plot we compare the range of the prediction performance spanned by the different RFs with other prediction approaches: it is evident that all variations of RF outperform the other models.\\
	The results of this analysis are:
	\begin{enumerate}
		\item \textit{min samples leaf} tuning brings to better predictions, but \textit{max depth} optimization speeds up the training time;
		\item Prediction performance is higher when using a low number of blocks (for instance, using BRIM with 8 blocks or no partition at all). We can deduce that the RF is, in a sense, able to recognize the blocks on its own. However, by using the BRIM blocks we can reach the same prediction power in less time;
		\item The more blocks we define with the community detection models, the more the RF is trained quickly, however, the performance tends to decrease;
		\item The 2 digit aggregation represent a bad definition of relatedness, since the RF trained with these blocks is the one that performs worse; for instance, IBN defines almost the same number of blocks and requires about the same computational time, but performs significantly better.
		\item Also the 21 HS sections do not represent a good definition of relatedness. This can be seen by comparing the performance with using $BRIM^2$, which provides the same performance with more blocks (42) and so less computational time;
		\item Even if we consider the worst model, that is the 2 digit blocks and the optimization on \textit{max depth}, the best F1 score is significantly better than the one of the network models.
	\end{enumerate}
	
	Now we motivate these results by investigating how these choices influence the training of the RF.\\	
	Result 1: \textit{max depth} speeds up the training time because it represents a more drastic constraint than \textit{min samples leaf}. Indeed, the average depth of a tree without any constraint is about 60, while the optimal value of \textit{max depth} we find is usually less than 10 (it is 15 only if we do not use blocks), and shallower trees are trained faster. On the other hand, \textit{min samples leaf} is a less drastic cut: what changes is at the level of the leaf nodes, so it is targeted to the removal only of the splits that bring to overfitting, and for this reason the performance is better. We can imagine that we have a real tree with some sick leaves and we have to remove the sick leaves knowing that the probability to have a sick leaf is proportional to the length of the branch. What we can do is either to cut all the branches longer than a certain threshold or to remove the sick leaves one by one. The first option corresponds to a \textit{max depth} tuning, it is faster, but it reduces the quality of the tree. The second option corresponds to a \textit{min samples leaf} tuning, it requires more time, but the quality of the resulting tree is better.\\
	Results 2 and 3: since the RF is able to recognize the blocks on its own, if we provide a good partition of the products what we can obtain is, at most, that we do not decrease its predictive power; however, if the selected partition contains too many blocks, we reduce the information the RF can learn, since each model has fewer products to see, and in this way we have a decrease in performance. However, with a higher number of blocks the training is faster for two reasons: the first is that the input has less features and the second is that, without a lot of products that have nothing to do with what we want to predict, the decision trees need less cuts and a lower depth.\\
	Results 4 and 5: let us consider the jewelry firm we discussed in section \ref{res1}. As we can see from table \ref{tab: gioielli}, its products are spread into different HS sections and chapters. So the HS does not provide good partitions for relatedness analyses at firm-level.\\
	Result 6: this result has practical consequences: if one wants to speed up the training of the algorithm through the use of (possibly good) partitions and a \textit{max depth} optimization, in any case, the RF will outperform all network models. Note, however, that realistic applications do not usually require real time investigations. 
	\section{Conclusions}
	The concept of relatedness, or coherence, is usually applied to quantify the closeness between an economic actor such as a country or a firm and an activity such as competitively exporting a given product. The possible practical applications of relatedness assessments are widespread; for instance, policymakers and institutions may want to quantify how much a developing country is far from entering into a given market (given its present export diversification) before deciding on an investment strategy, or if a new product is feasible given the present export basket of a firm. In both the mainstream and in the economic complexity literature, relatedness is measured in two steps. First of all, one builds a network of economic activities, typically products, in which the weights of the links are given by the so-called co-occurrences: the more countries export both products, the more the two will be similar. Different ways of building such networks co-exist in the literature. The second step consists in computing the relatedness as the average similarity between the exports of a given country and the target product. In this paper, we discuss and investigate two radical improvements: the use of supervised machine learning and firm-level, instead of country-level, data. In order to quantitatively compare the resulting different measures of relatedness, we test them against a forecast task, the assumption being that, on average, an economic actor will likely diversify in products that are relatively more related to. By means of both specific examples and general statistical assessments, we are able to show that: i) machine learning, and in particular random forest, outperforms network-based methods regardless of the data typology; ii) firm-level data provides a better assessment of relatedness, in the sense that while a model built on country-level data is totally unsuitable to predict future exports of firms, a model built on firm-level data is still able to accurately predict future exports of countries. This is due to the relative specialization of firms, that accurately tracks the similarity between products. On the contrary, successful countries are highly diversified, providing misleading co-occurrences; iii) community detection algorithms provide partitions in sub-sets of products which reduce the computational effort needed to train the algorithms, iv) regardless of the method used to build the relatedness measure, the optimal strategy is to train the forecast model using data of the same typology one wants to predict (in particular, firm-level data to build relatedness measures to be used at firm level).\\
	In summary, in order to compute the feasibility of a product for a firm, one should use machine learning algorithms trained on firm-level data, since the widespread use of co-occurrences computed at country level leads to poor assessments of the relatedness.  \\
	These results open up a number of consequent investigations. First, the very same exercise should be replicated with different kinds of human activities, for instance, patents. Indeed, the relatedness between technological sectors is usually measured by counting country-level co-occurrences, and this assessment very likely suffers from the same issues which we have exposed, and solved, here. Second, relatedness enters in a number of derived quantities which are used to characterize the diversification strategies of countries, firms, and regions. The robustness of these quantities should be checked in light of the findings hereby reported. Finally, the validation strategy we propose to quantitatively compare the different relatedness measures - a rigorous out-of-sample forecast exercise - could be applied, more in general, to the various concepts used in economic complexity, in order to scientifically validate or falsify the different approaches, an issue of general relevance in the physics of complex systems.

	\section{Declaration of Competing Interest}
	The authors declare that they have no known competing financial interests or personal relationships that could have appeared to influence the work reported in this paper.
	\section{Acknowledgments}
	This work was supported by the Centro Ricerche Enrico Fermi research project "complessità in economia".
	We thank ISTAT for providing the Italian firms data in particular Dr. Stefano Menghinello and Dr.	Cristina Lanzi, we thank Francis Farrelly for data encryption and pre-processing, and Luciano Pietronero for useful discussions. An earlier version (preprint) of this work can be found in https://arxiv.org/abs/2202.00458.  
	
	\bibliographystyle{vancouver}
	\bibliography{biblio}
	
\end{document}